\documentclass[10pt,twocolumn,letterpaper]{article}

\usepackage[pagenumbers]{cvpr}
\usepackage{graphicx}
\usepackage{amsmath}
\usepackage{amssymb}
\usepackage{booktabs}
\usepackage{microtype}
\usepackage{enumitem}
\usepackage{array,multirow,textcomp}
\usepackage{balance}

\renewcommand{\paragraph}[1]{%
    \vspace{0.5\baselineskip}
    \noindent\textbf{#1}
    \hspace{0.5em}
}

\newcolumntype{Y}{>{\centering\arraybackslash}X}
\newcolumntype{x}[1]{>{\centering\arraybackslash\hspace{0pt}}p{#1}}
\newcolumntype{C}[1]{>{\centering\let\newline\\\arraybackslash\hspace{0pt}}m{#1}}
\newcolumntype{R}[2]{%
    >{\adjustbox{angle=#1,lap=\width-(#2)}\bgroup}%
    l%
    <{\egroup}%
}

\newcommand{\spm}[1]{\tiny{$\,\pm$#1}}

\newcommand{\fixedvspace}[1]{%
  \par\kern-\prevdepth\vspace{#1}%
}

\usepackage[normalem]{ulem}
\usepackage[pagebackref,breaklinks,colorlinks]{hyperref}

\usepackage[capitalize]{cleveref}
\crefname{section}{Sec.}{Secs.}
\Crefname{section}{Section}{Sections}
\Crefname{table}{Table}{Tables}
\crefname{table}{Tab.}{Tabs.}

\def\cvprPaperID{25}
\def\confName{ICCV}
\def\confYear{2023}

\begin{document}

\title{
Breathing New Life into 3D Assets with Generative Repainting
}

\author{%
\!\!
Tianfu Wang\textsuperscript{\textnormal{1}}\quad\ 
Menelaos Kanakis\textsuperscript{\textnormal{1}}\quad\ 
Konrad Schindler\textsuperscript{\textnormal{2}}\quad\ 
Luc Van Gool\textsuperscript{\textnormal{1,3,4}}\quad\ 
Anton Obukhov\textsuperscript{\textnormal{1\raisebox{0.3ex}{\tiny$\rightarrow$}2}}\\
{\small
\textsuperscript{\textnormal{1}}ETH Z\"urich, Computer Vision Laboratory\quad\ \ 
\textsuperscript{\textnormal{2}}ETH Z\"urich, Photogrammetry and Remote Sensing\quad\ \ 
\textsuperscript{\textnormal{3}}KU Leuven\quad\ \ 
\textsuperscript{\textnormal{4}}INSAIT, Sofia
}
}

\twocolumn[{%
\renewcommand\twocolumn[1][]{#1}%
\maketitle
\begin{center}
  \newcommand{\teaserwidth}{1.0\textwidth}
  \centerline{
    \includegraphics[width=\teaserwidth]{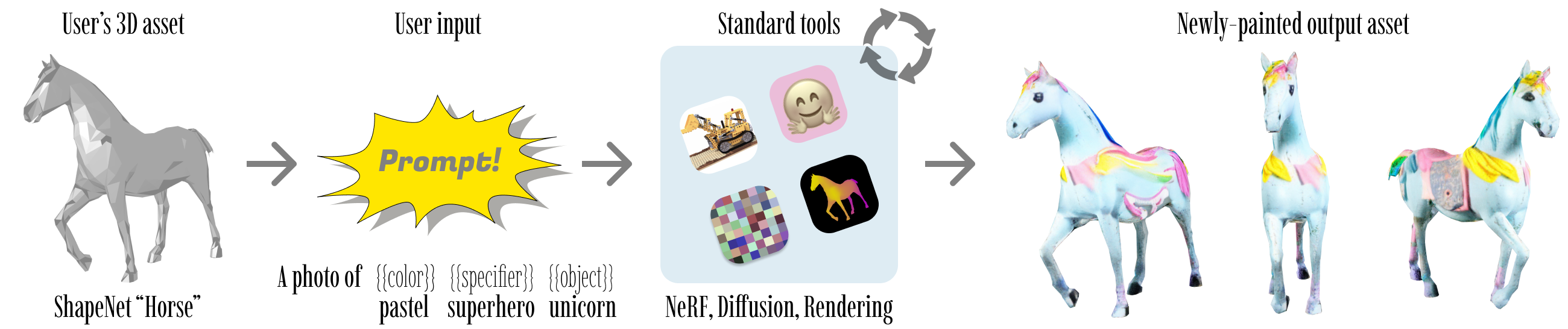}
  }
  \captionsetup{type=figure}
  \captionof{figure}{
    \textbf{We present a pipeline for text-guided painting of legacy geometry.}
    We leverage rich pretrained generative 2D diffusion models to give a fresh look to existing 3D assets, and neural radiance fields to enforce 3D consistency and overcome issues of the legacy representations.
    Starting from an input geometry and the desired output description, our pipeline orchestrates calls to several generative and modality conversion tools to breathe new life into the input assets.
    The tools communicate using images instead of gradients with each other, making our pipeline interpretable and amenable to partial upgrades.
    Project page: \href{https://www.obukhov.ai/repainting_3d_assets}{https://www.obukhov.ai/repainting\_3d\_assets}.
  }
  \label{fig:teaser}
 \end{center}%
}]

\begin{abstract}
Diffusion-based text-to-image models ignited immense attention from the vision community, artists, and content creators. 
Broad adoption of these models is due to significant improvement in the quality of generations and efficient conditioning on various modalities, not just text. 
However, lifting the rich generative priors of these 2D models into 3D is challenging.
Recent works have proposed various pipelines powered by the entanglement of diffusion models and neural fields.
We explore the power of pretrained 2D diffusion models and standard 3D neural radiance fields as independent, standalone tools and demonstrate their ability to work together in a non-learned fashion. 
Such modularity has the intrinsic advantage of eased partial upgrades, which became an important property in such a fast-paced domain.
Our pipeline accepts any legacy renderable geometry, such as textured or untextured meshes, orchestrates the interaction between 2D generative refinement and 3D consistency enforcement tools, and outputs a painted input geometry in several formats.
We conduct a large-scale study on a wide range of objects and categories from the ShapeNetSem dataset and demonstrate the advantages of our approach, both qualitatively and quantitatively.
\end{abstract}

\section{Introduction}
\label{sec:intro}
Creating high-quality 3D assets based on textual descriptions for a diverse range of objects is an endeavor with great potential for digital media and artists. 
Recently, there has been a rise in denoising diffusion-based (DDPM)~\cite{ddpm} text-to-image models~\cite{saharia2022photorealistic, Rombach_2022_CVPR} producing results of unprecedented quality. 
The generative power of these 2D image models prompts the question: Can we use them to generate multi-view consistent 3D content?
As it turns out, lifting these rich generative priors to 3D is a non-trivial task. 
In this work, we focus on the problem of text- and geometry-conditioned painting, an adjacent problem of text-to-3D generation.

\begin{figure*}[t!]
  \centering
  \includegraphics[width=1\textwidth, trim={5em 0 4em 0}, clip]{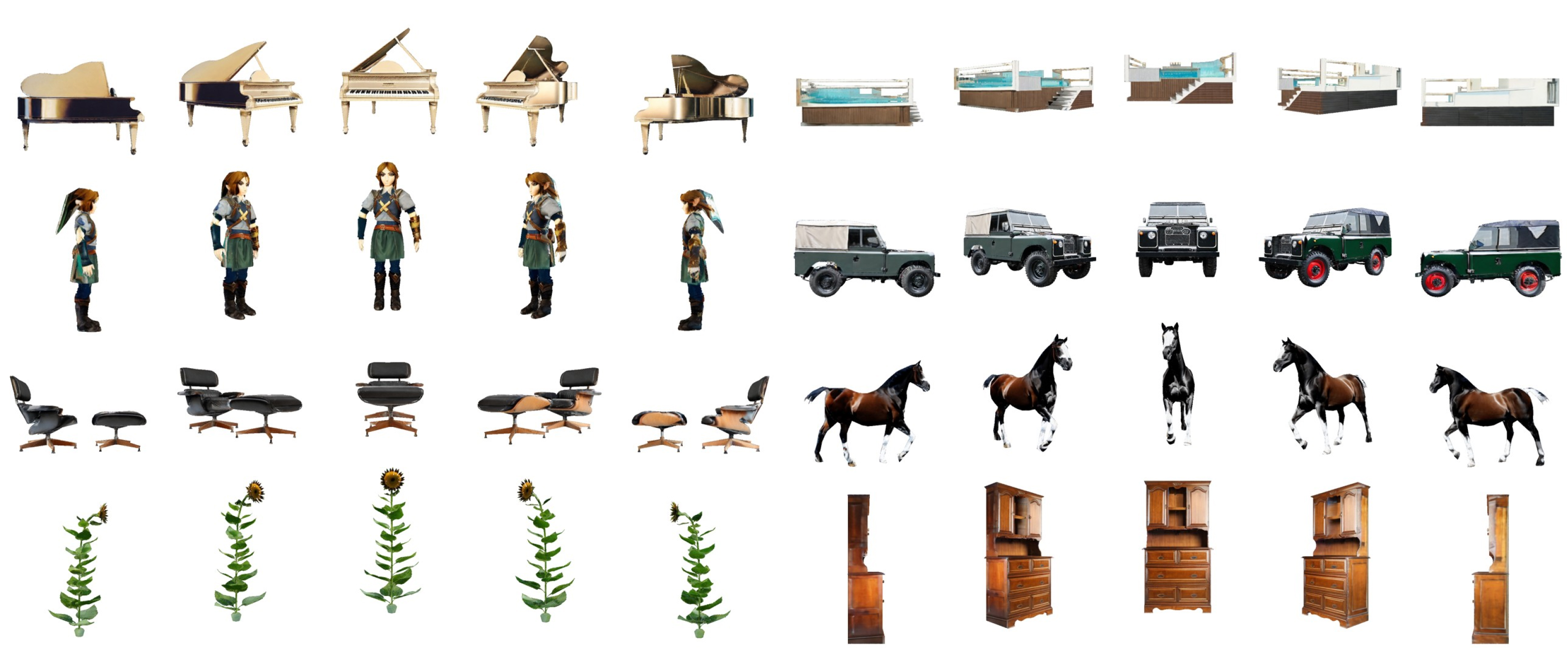}
  \caption{
    \textbf{Texturing the ShapeNetSem~\cite{chang2015shapenet} dataset with the proposed method}. 
    We discard the original texture and paint objects with our method using the dataset metadata ``name'' field as a text prompt.
    We show several objects from 5 views spaced with 45-degree increments around the vertical axis.
    Our method produces high-quality results from the input text and geometry.
    More visual results in Figs.~\ref{fig:comparison:big},~\ref{fig:shapenetsem:youtube}.
  }
  \label{fig:centerpiece}
\end{figure*}

The overview of our pipeline for generating a diverse multi-view consistent painting from a text description and input geometry is presented in Fig.~\ref{fig:teaser}. 
We bootstrap our pipeline from two crucial components: a pretrained generative text- and depth-conditioned image diffusion model~\cite{Rombach_2022_CVPR} and neural radiance fields (NeRF)~\cite{mildenhall2021nerf}.
The design of our pipeline separates these components into distinct processes, which communicate using the interface of image files.
This is contrary to several recent approaches that rely on gradient flow between the components, either in the form of Score Distillation~\cite{poole2022dreamfusion, metzer2022latent}, or differentiable rendering~\cite{richardson2023texture}.
We rely on traditional rendering techniques to enable communication between the components, including Z-buffer extraction for the rendered views.
The image file interface is naturally interpretable and better suited for building modular and partially upgradable systems.
This is especially important as both DDPM and NeRF research fields advance rapidly.

Prior generative 3D works often employ the UV texture unwrapping~\cite{levy2002least}, a costly operation and a potential point of failure.
Since our method requires only Z-buffer queries from the input geometry, the input does not necessarily have to have a UV texture map attached or even be a valid mesh.
To support this claim, we experimented with \mbox{Point-E}~\cite{nichol2022point}, thus extending our pipeline to a pure text-to-3D setting.

The output of our pipeline is a NeRF corresponding to the input geometry, painted in a multi-view consistent manner.
The NeRF can be converted into the explicit input format with extra coloring information.

Our pipeline's performance depends on each component's performance, so it will keep improving as the components get faster. 
For example, recent progress in DDPMs~\cite{Rombach_2022_CVPR} led to a tenfold decrease in image generation time; NeRF research has seen similar speedups~\cite{muller2022instant}. 
Capitalizing on that, we conduct a large-scale study of painting the ShapeNetSem~\cite{chang2015shapenet} dataset, composed of 12K objects from over 270 categories (Fig.~\ref{fig:centerpiece}).
The study shows that our pipeline sets new state-of-the-art results on several generative metrics while attaining proper 3D consistency.

The summary of our contributions is as follows:
\begin{itemize}
  \item We introduce a novel approach for giving 3D assets a new life, by painting their geometry using text inputs and pretrained generative image diffusion models.
  \item Our method is unique in that it combines pretrained 2D diffusion models and 3D neural radiance fields as \textit{standalone} pipelines. 
  The weak coupling of tools is achieved through the interpretable interface of image files and permits partial upgrades.
  \item We conduct a large-scale study of painting ShapeNetSem~\cite{chang2015shapenet} dataset and attain the new state-of-the-art on several metrics and perceived 3D consistency.
  \item Our method is robust to input corruptions and produces the output assets in several formats. 
\end{itemize}

\section{Related Work}

\begin{figure*}[t!]
    \centering
    \includegraphics[width=1.0\textwidth]{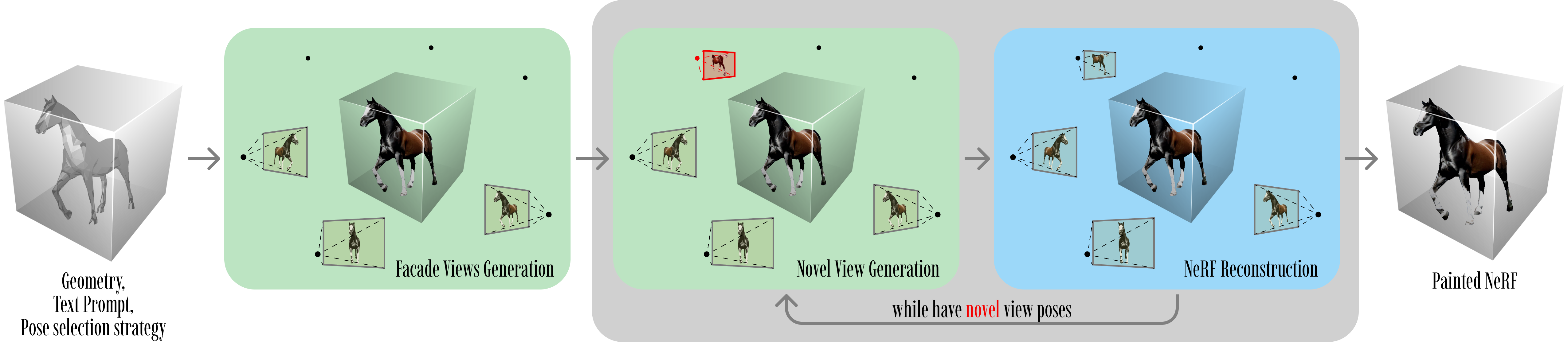}
    \caption{
      \textbf{Geometry painting pipeline that takes the geometry, a text prompt, and outputs a painted NeRF of the model.} 
      We utilize the diffusion image generation process and the 3D reconstruction process of NeRF as standalone procedures. 
      We start by generating the facade views using only diffusion view generation. 
      Our pipeline progressively builds the 3D model by using NeRF to generate view-consistent images and feeding them back to the diffusion process to generate a new input view according to the view selection strategy. 
    }
    \label{fig:overview}
\end{figure*}

\paragraph{Generative Text-to-Image Models}
Until recently, generative imaging was dominated by unconditional or few-classes-conditional models~\cite{goodfellow2014generative,sohl2015deep,cai2022pix2nerf}.
With advancements in natural language processing, Contrastive Language-Image Pretraining (CLIP)~\cite{radford2021learning} bridged the gap between visual and text modalities.
This opened an avenue for open-category and text-conditioned image generation.
Currently, Denoising Diffusion Probabilistic Models (DDPM)~\cite{ddpm,saharia2022photorealistic} dominate the niche of high-quality and affordable text-conditioned generative imaging.
Stable Diffusion~\cite{Rombach_2022_CVPR} proposed shifting the diffusion process to a low dimensional latent space, achieving competitive performance while reducing the computation requirements. 
Subsequent models could further condition the process on various modalities, such as depth maps, images, and inpainting masks.
These new modalities and accessible pretrained checkpoints gave rise to new applications of diffusion models, such as image un-cropping~\cite{saharia2022palette} and perpetual view generation~\cite{cai2022diffdreamer}. 
Likewise, our method relies on standalone pretrained DDPMs with their various ways of conditioning.

\paragraph{Neural Radiance Fields}
Neural scene representations gained popularity due to their simplicity of usage and ability to capture complex scenes efficiently. 
Neural Radiance Fields (NeRF)~\cite{mildenhall2021nerf} have recently demonstrated their versatility as a solution for 3D reconstruction from posed images.
Recently, numerous improvements and variants of NeRF have been developed~\cite{muller2022instant,chen2022tensorf,obukhov2022tt}. 
In particular, Instant NGP~\cite{muller2022instant} proposed an efficient multi-resolution 
hash-based 
grid data structure, which reduces the training time of NeRF from hours to minutes.
Similarly to COLMAP~\cite{schoenberger2016sfm} for structure for motion, Instant NGP has become the go-to standalone tool for images to NeRF conversion.

\paragraph{Generative 3D Models}
Research on high-quality 3D models and assets generation gained a lot of interest recently~\cite{gao2022get3d, siddiqui2022texturify, wu2016learning, poole2022dreamfusion, watson2022novel}. 
Previous methods leveraged Generative Adversarial Networks (GANs)~\cite{goodfellow2014generative} coupled with 3D-aware learned pipelines, such as differentiable renderers~\cite{gao2022get3d}, face convolutional neural networks (CNNs)~\cite{siddiqui2022texturify}, voxel grids~\cite{wu2016learning}, and NeRFs~\cite{xu20223d, cai2022pix2nerf}. 
However, most of the methods require training a separate model per category, and thus, the evaluation focuses on a handful of classes, typically ``cars'' and ``chairs'', such as seen in ShapeNet~\cite{chang2015shapenet}.
With the rise of popularity in diffusion models and accessible text conditioning, recent works focused on integrating them into 3D content generation pipelines~\cite{poole2022dreamfusion, watson2022novel}. 
DreamFusion~\cite{poole2022dreamfusion} proposed score distillation sampling to couple a pretrained text-to-image diffusion model with a NeRF module to form an end-to-end trainable pipeline. 
Although score distillation cleverly avoids backpropagation through the diffusion model, thus reducing computational costs, it still requires significant computations. 
Pipelines with surrogate 3D output~\cite{nichol2022point} have also received attention.
Mesh-based inpainting schemes such as Latent-Paint~\cite{metzer2022latent} and TEXTure~\cite{richardson2023texture} employ differentiable rendering to generate a texture image for the input mesh. 
However, these methods are susceptible to artifacts introduced during UV texture unwrapping and gradient interaction between the generative model and the texturing target. 
Another two relevant works appeared recently: Text2Tex~\cite{chen2023text2tex} utilizes a mesh-based inpainting scheme similar to TEXTure~\cite{richardson2023texture}; TextMesh~\cite{tsalicoglou2023textmesh} combines NeRF with SDS loss akin to DreamFusion~\cite{poole2022dreamfusion}.
Our method overcomes the discussed limitations by using NeRF for both scene representation and iterative consistency enforcement. 

\section{Method}
\label{sec:method}

The pipeline of our method is outlined in Fig.~\ref{fig:overview}.
It takes an input geometry and a text description and generates a NeRF model that adheres to the structure of input geometry but is enhanced with text-guided painting. 
It paints the geometry progressively: starting from the object facade initialization, it iteratively picks a novel view according to the camera pose selection strategy, generates a novel view, and reconciles it with the previous views using NeRF.

\paragraph{Prerequisites and Assumptions}
Our pipeline is object-centric; hence, we create a virtual scene with the object scaled and positioned in the origin and a camera positioned on a unit sphere, pointing at the origin. 
We additionally assume that the object surface is opaque, which is required to perform unambiguous queries of the renderer's Z-buffer. 
This constraint limits processing models with transparency or with large sprite surfaces (e.g., for trees or flowers), sometimes seen in ShapeNetSem.
As discussed in the previous chapters, the input geometry is not required to have UV unwrapping or other properties attached to the geometry.
Whenever normals are available, the inpainting procedure can benefit from them through an additional inpainting zoning step; however, this is optional.

We require a pretrained image diffusion model with text and depth conditioning to paint novel views. 
From the NeRF pipeline, we expect that it can ingest view images, poses, and optional depth maps, and output a model that can be queried at arbitrary poses for color and depth.

\begin{figure*}[t!]
    \centering
    \includegraphics[width=\textwidth]{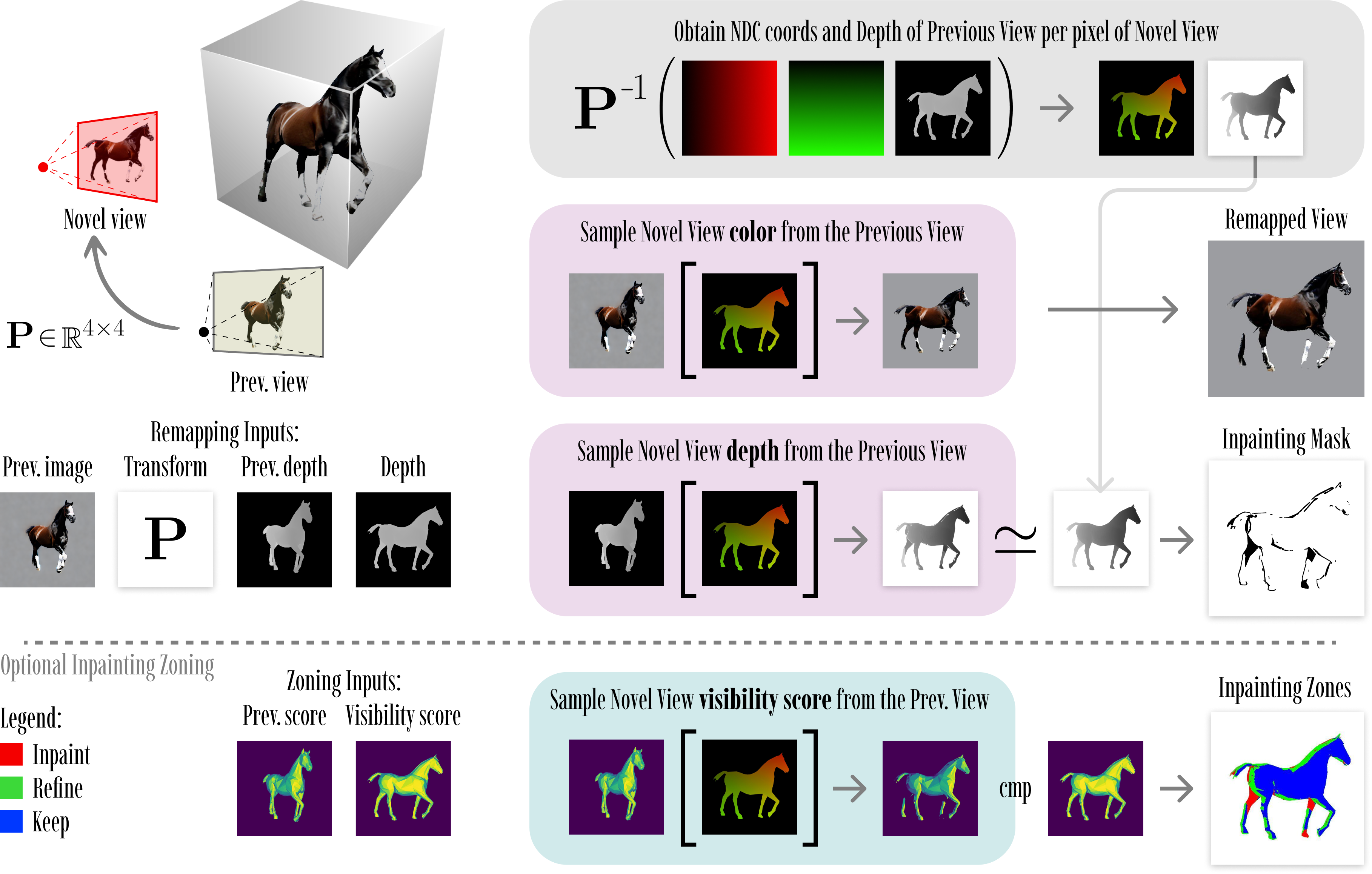}
    \caption{
    \textbf{
        Novel view remapping from a previous view.    
    }
    Multi-view consistency is enforced by remapping the previous view into the novel view and preparing the inpainting mask of the unseen areas.
    The remapping procedure consists of three steps: 
    (1) obtaining sampling coordinates of the previous view in the novel view, 
    (2) sampling the novel view from the previous view, and 
    (3) obtaining the inpainting mask by analyzing occlusions. 
    An optional inpainting zoning step provides better control of inpainting for inputs with surface normals.
    } 
    \label{fig:mapping}
\end{figure*}

\paragraph{Initialization}
The first view generation defines and constrains the object's overall painting and style. 
To obtain the first painted view, we render the object's depth map and give it together with the text prompt to the depth-to-image pipeline. 
At this point, it is possible to query the user if the generated initialization is according to expectation and make early alterations by changing text or the pipeline seed.

\paragraph{Novel View Remapping}
Multi-view consistency is crucial for generating meaningful geometry painting. 
However, it is tricky to achieve in a pipeline with disentangled stages applied one after another, such as our design. 
To this end, we employ an occlusion-aware backward remapping scheme for image view reprojection from a previously-painted view to the novel one (Fig.~\ref{fig:mapping}). 

At its core is the view transformation $\mathbf{P}=\mathbf{K} \mathbf{E} \mathbf{K}^{-1}$, which transforms normalized device coordinates (NDC) of the previous view into the novel view, where $\mathbf{K}$ is the projection from world to NDC space and $\mathbf{E}=[\mathbf{R} | \mathbf{T}]$ is the relative transformation of camera poses in world coordinates.

\begin{figure*}[t!]
    \centering
    \includegraphics[width=\linewidth]{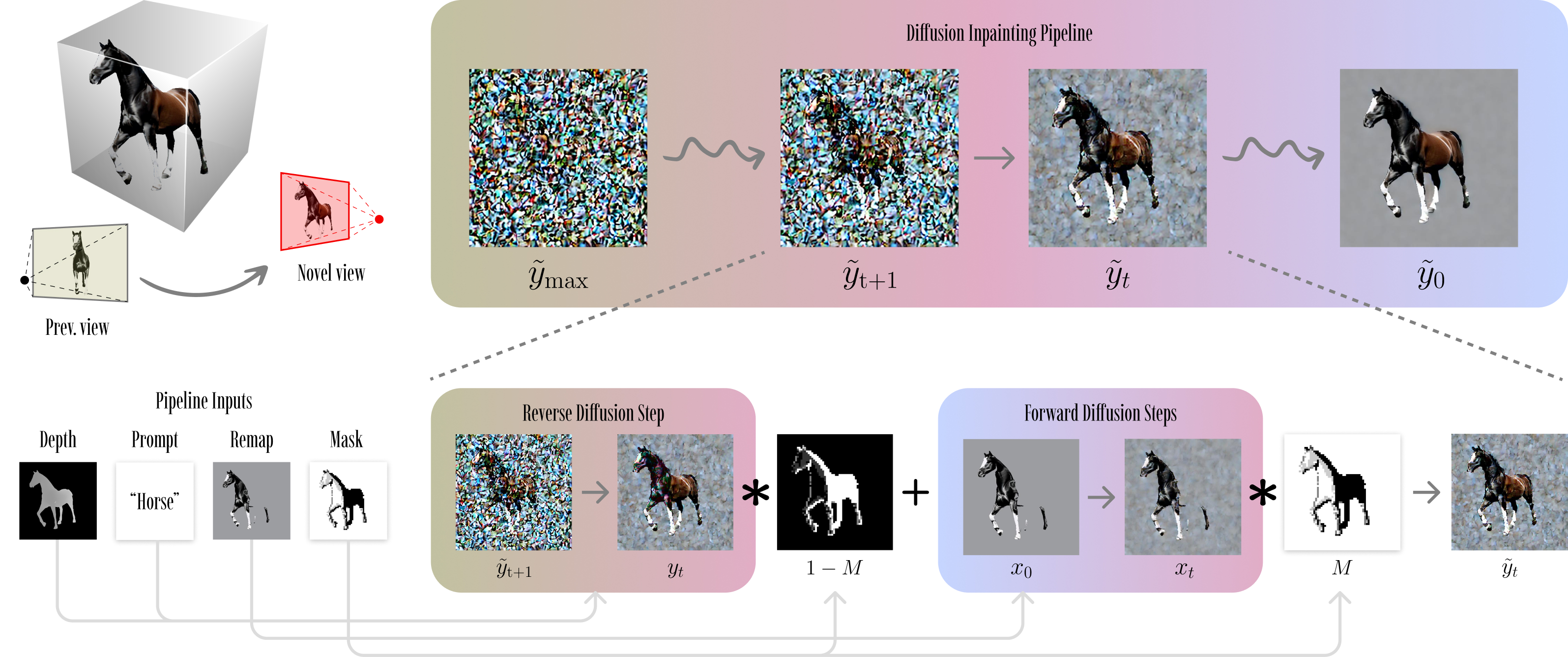}
    \caption{  
    \textbf{
    Our text- and depth-conditioned latent diffusion inpainting pipeline for constrained novel view synthesis.
    }
    It is inspired by both the inpainting pipeline that takes an inpainting mask and applies it to the latents, and the text- and depth-conditioned generation pipeline from the Stable Diffusion distribution~\cite{Rombach_2022_CVPR}.
    At each diffusion time step, the latents are composed from the forward diffusion step over the inpainting constraints (``Remap'' in the figure), and the reverse diffusion step, conditioned on the input text prompt and depth. 
    }
    \label{fig:diffusionviewgen}
\end{figure*}

As a first step, we use the inverse transform $\mathbf{P}^{-1}$ to map the novel view NDC coordinates with $z$-values assigned from the Z-buffer of the novel view rendering into the previous view. 
This gives us an $xy$-map (depicted as a green-red tile) of pixels of the novel view and their source locations directly in the previous view. 
The transform also gives us the depth map of the source locations as seen from the previous view, which is used for the occlusion test.

Secondly, we obtain the previous view's backward remapping into the novel view using the bilinear interpolation of the previous view at the $xy$-map locations. 
Compared to the direct application of the transform $\mathbf{P}$ to the previous view image, the backward remapping is continuous by design and guarantees the absence of seams or holes in the remapped image. 

However, an additional occlusion mask is required to identify areas of the novel view that are not visible from the previous view to handle these areas properly. 
Thus, as a third step, we obtain this mask by comparing the previous view depth map resampled using our $xy$-map, with the $z$-values obtained from the transformation on the first step.
Evidently, the positions with agreeing depth are visible in both views under the assumptions we declared in the prerequisites.
The final remapped view is thus obtained by combining the outputs of the previous two steps.

Additionally, the occlusion mask is stored for the future inpainting stage. 
Since most inpainting methods permit varying inpainting strength per pixel, we additionally compute inpainting zones map (similar to ``trimaps'' in TEXTure~\cite{richardson2023texture}), whenever the input geometry has surface normals.
Specifically, we assign the visibility score to each fragment as a dot product between the surface normal and the unit vector originating in the camera origin and pointing at the fragment.
By comparing visibility scores between the previous and novel views' fragments, we classify zones into areas that are kept intact, areas of full inpainting, or refinement.
As we identify in the ablation study, inpainting zoning helps with multi-view consistent painting details.

Finally, as we expand the painted area of the input, more views become available for color transfer to a novel view. 
The described procedure is thus easily extended to perform remapping from multiple previous views. 

\paragraph{Novel View Inpainting}
We employ a custom text- and depth-conditioned latent diffusion inpainting pipeline to complete novel views after the remapping.
The pipeline inherits from the previous works on inpainting with diffusion models~\cite{lugmayr2022repaint, couairon2022diffedit} and is largely based on the pretrained Stable Diffusion~\cite{Rombach_2022_CVPR}. 
The input to the pipeline is the same as for image generation, with the addition of a mask that defines the inpainting area and the remapped image constraint (Fig.~\ref{fig:diffusionviewgen}). 

The mask $M$ is taken from the remapping stage and downsampled to match the latent diffusion resolution.
Upon availability, inpainting zoning additionally assigns an intermediate weight value for the refined areas. 

At each denoising step $t$, we take the latent representation of the remapped image $x_0$ and inject noise through $t$ forward diffusion steps to obtain $x_t$.
At the same time, we perform a single reverse diffusion step to obtain $y_t$ from the more noisy $\tilde{y}_{t+1}$ step, at which point we use the depth and the text prompt as conditions.
We now blend the denoised latent $y_t$ with remapped conditoon $x_t$ using the inpainting mask $M$: $\tilde{y_t} = (1 - M)y_t + M x_t$.
This process starts with $\tilde{y}_\mathrm{max} \sim \mathcal{N}(0,1)$ and is repeated until obtaining $\tilde{y}_0$, which is then decoded into the inpainting output.
Notably, latent diffusion is the primary source of inconsistency between the inpainted images and their remapped constraints, which calls for a solution to enforce multi-view consistency globally. 

\begin{figure*}[t!]
    \centering
    \includegraphics[width=\linewidth]{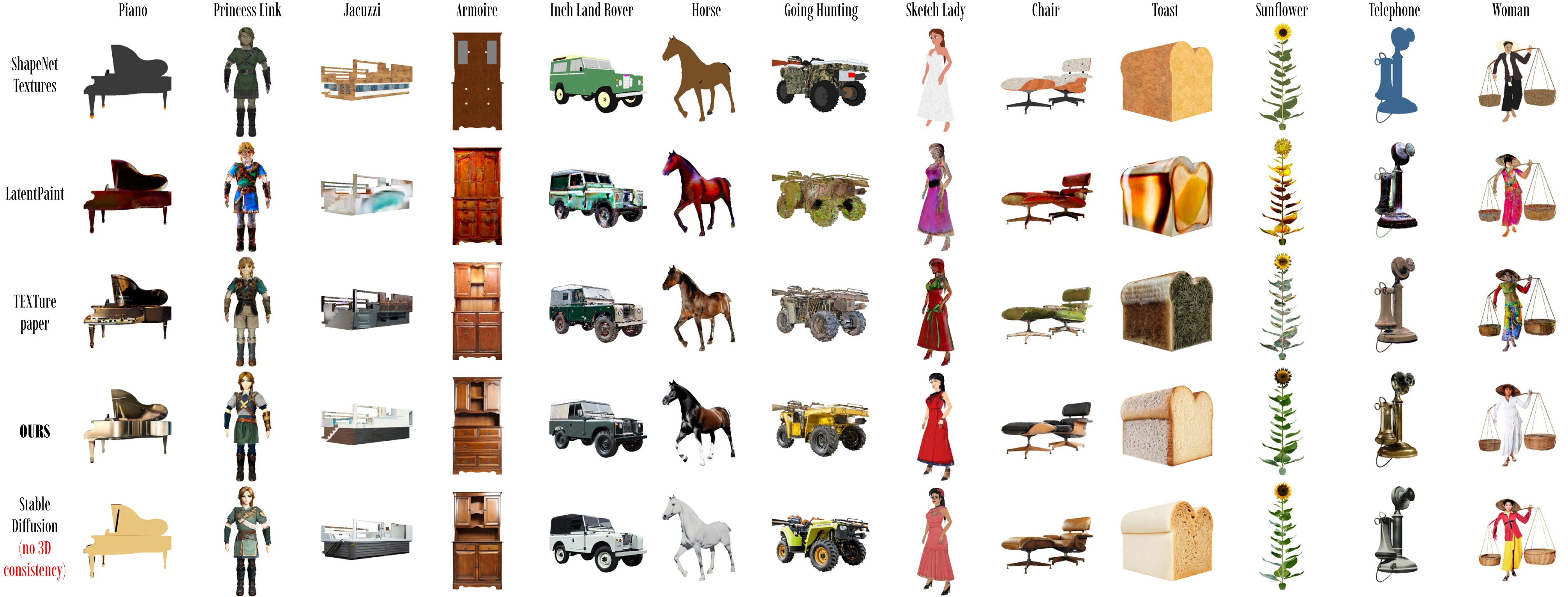}
    \caption{
      \textbf{Qualitative Comparisons} of our method to TEXTure~\cite{richardson2023texture}, Latent-Paint~\cite{metzer2022latent}, the original texturing from ShapeNetSem~\cite{chang2015shapenet}, and the ``upper-bound quality'' generative prior applied to each individual view without 3D consistency constraints.
      As can be seen, our method generates noise- and seam-free texturing with a high degree of detail. 
    }
    \label{fig:comparison:big}
\end{figure*}

\paragraph{NeRF Reconstruction}
Using the remapping and inpainting techniques introduced above, we can ensure the soft consistency of a subset of proximal views. 
However, we aim for global multi-view consistency, which requires considering all the generated views simultaneously.
To this end, we employ a flavor of NeRF to resolve multi-view conflicts and reconcile painting from all viewpoints.
Since the standard NeRF formulation supports different colors of the same 3D location depending on the viewpoint, we disable such view-dependent effects and fit the NeRF to predict view-invariant colors instead.
Starting with a set of facade views and until there are no more unvisited poses, we submit all the generated images, their respective camera poses, and depth maps, as inputs to NeRF.
Once the scene is fitted, all painted training images are replaced with renders from the fitted NeRF, so that our subsequent remapping steps always start from multi-view consistent inputs.

\section{Experiments}
\label{sec:exp}

As a first step towards painting ShapeNetSem, we chose a few hyperparameters for our pipeline.
To paint each model, we rely on 9 views regularly spaced around the object in the horizontal plane ($40^\circ$ increment).
Starting from the front view, we generate 5 facade views using just the remapping and inpainting procedures.
This facade configuration maximizes the coverage of the input geometry within the range of efficiency of our remapping technique.
Before generating each subsequent view, we perform NeRF reconstruction.
Our pose selection strategy picks the next view from the clockwise and counter-clockwise increments in alternating steps.
We remap two of the closest painted views from the left and right paths around the model each time.
This technique helps minimize the content gap in the last view, where the clockwise and counter-clockwise painting paths meet.

\paragraph{Text Prompting}
The base is set to ``\textit{A photo of \{\!\{object\}\!\}}''.
An additional ``\textit{\{\!\{dir\}\!\} view}'' modifier specifies the coarse relation of the viewpoint and the object, helping with 3D consistency.
Other modifiers are discussed in the Appendix.

\paragraph{NeRF Setup}
We chose Instant NGP~\cite{muller2022instant} as a standalone NeRF backbone for its high degree of configurability and great performance.
Additionally, we leverage depth supervision in NeRF training to facilitate faster convergence and obtain higher-quality reconstruction.

Our setting slightly differs from the default NeRF objective because our training images are generated from diffusion and can have soft view conflicts. 
As mentioned previously, the purpose of NeRF in our pipeline is to bring multi-view painting to agreement rather than to simulate light transport.
We disable view-dependent effects in the NeRF configuration to align with this purpose.
Additionally, we adjust the parameters for the grid encoding settings. 
We found that a higher number of levels (5) and encoded features (16) achieve good rendering fidelity while keeping a sufficiently smooth and continuous NeRF surface.

\paragraph{ShapeNetSem Processing}
We demonstrate that our method can be applied to a wide range of object categories and shapes by conducting a study of texturing a significant subset of the ShapeNet~\cite{chang2015shapenet} dataset called ShapeNetSem, which contains 12K models in over 270 categories. 
We preprocess each model by orienting it using the up and front vectors from the metadata, centering, and scaling to fit the unit sphere.
We take the text prompt's ``\textit{object}'' part from the \texttt{name} field of the dataset metadata.

\begin{figure}[t!]
    \centering
    \href{https://www.youtube.com/watch?v=S-MUFPurJpc}{
      \begin{minipage}[t]{\linewidth}%
        \centering
        \includegraphics[width=\linewidth]{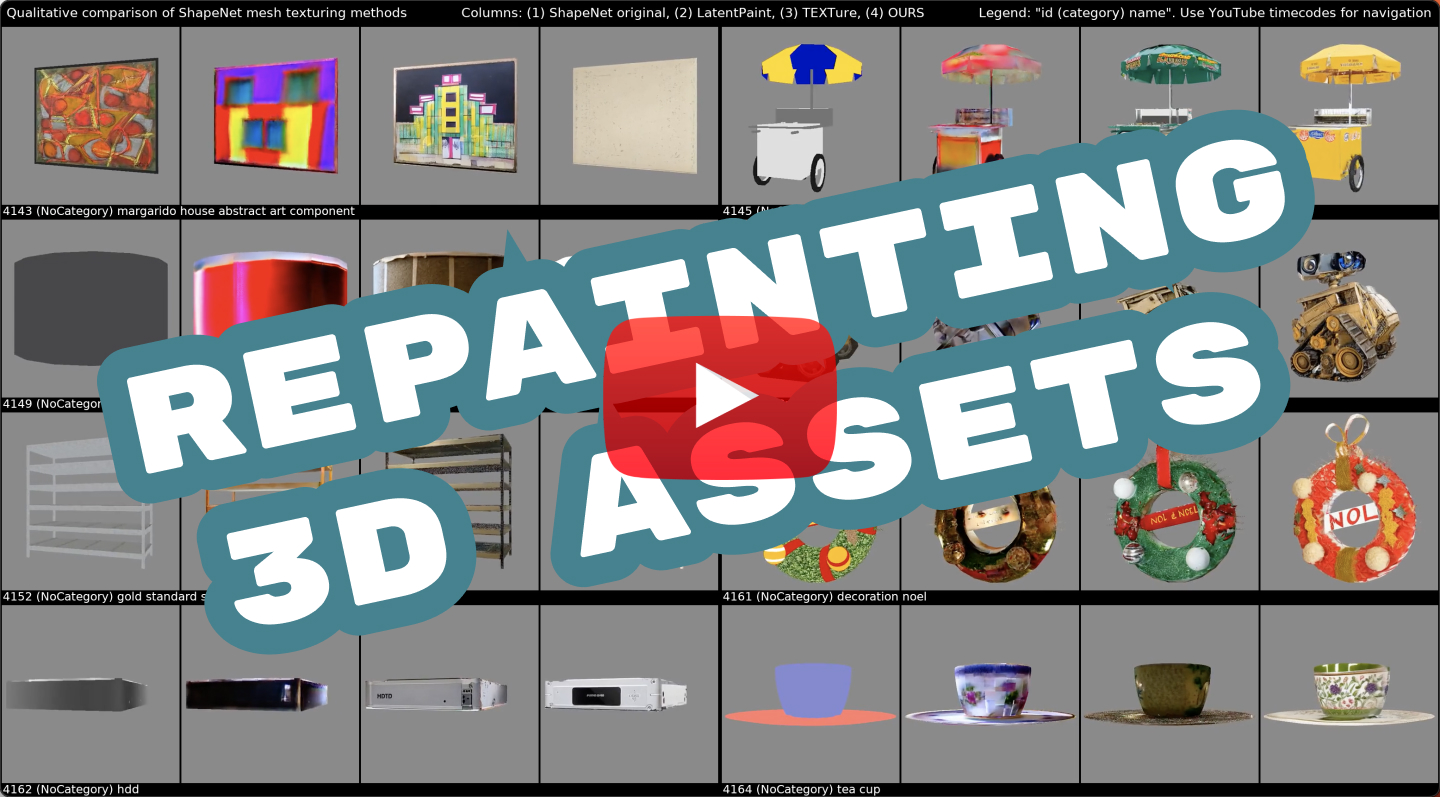}
        \smallskip
        \scriptsize{https://www.youtube.com/watch?v=S-MUFPurJpc}
      \end{minipage}%
    }%
    \caption{
      \textbf{Large-Scale Comparison of ShapeNetSem Texturing} with the original textures~\cite{chang2015shapenet}, Latent-Paint~\cite{metzer2022latent}, TEXTure~\cite{richardson2023texture}, and our method. 
      We present spin-views of $\sim$12K models from over 270 categories. 
      The models are grouped by category and sorted by group size.
      Categories, IDs, and model names (prompts) are specified under the corresponding video tiles.
      \textit{Tip}: Use timecodes to conveniently skip to categories of interest.
    }
    \label{fig:shapenetsem:youtube}
\end{figure}

\begin{figure}[t]
    \centering
    \includegraphics[width=1\linewidth]{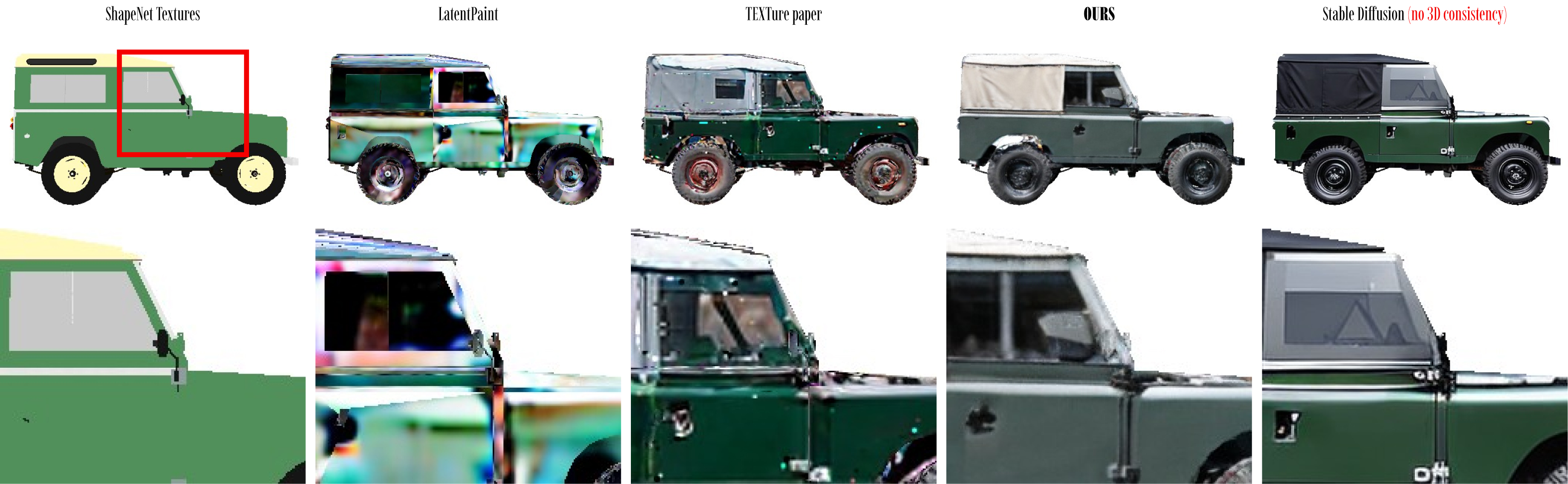}
    \caption{
      \textbf{A Closer Look} reveals that our method produces more realistic results with invisible seams, while other methods often exhibit texture filtering issues and lower realism.
    }
    \label{fig:comparison:zoomin}
\end{figure}

\begin{figure}[t]
  \centering
  \includegraphics[width=\linewidth]{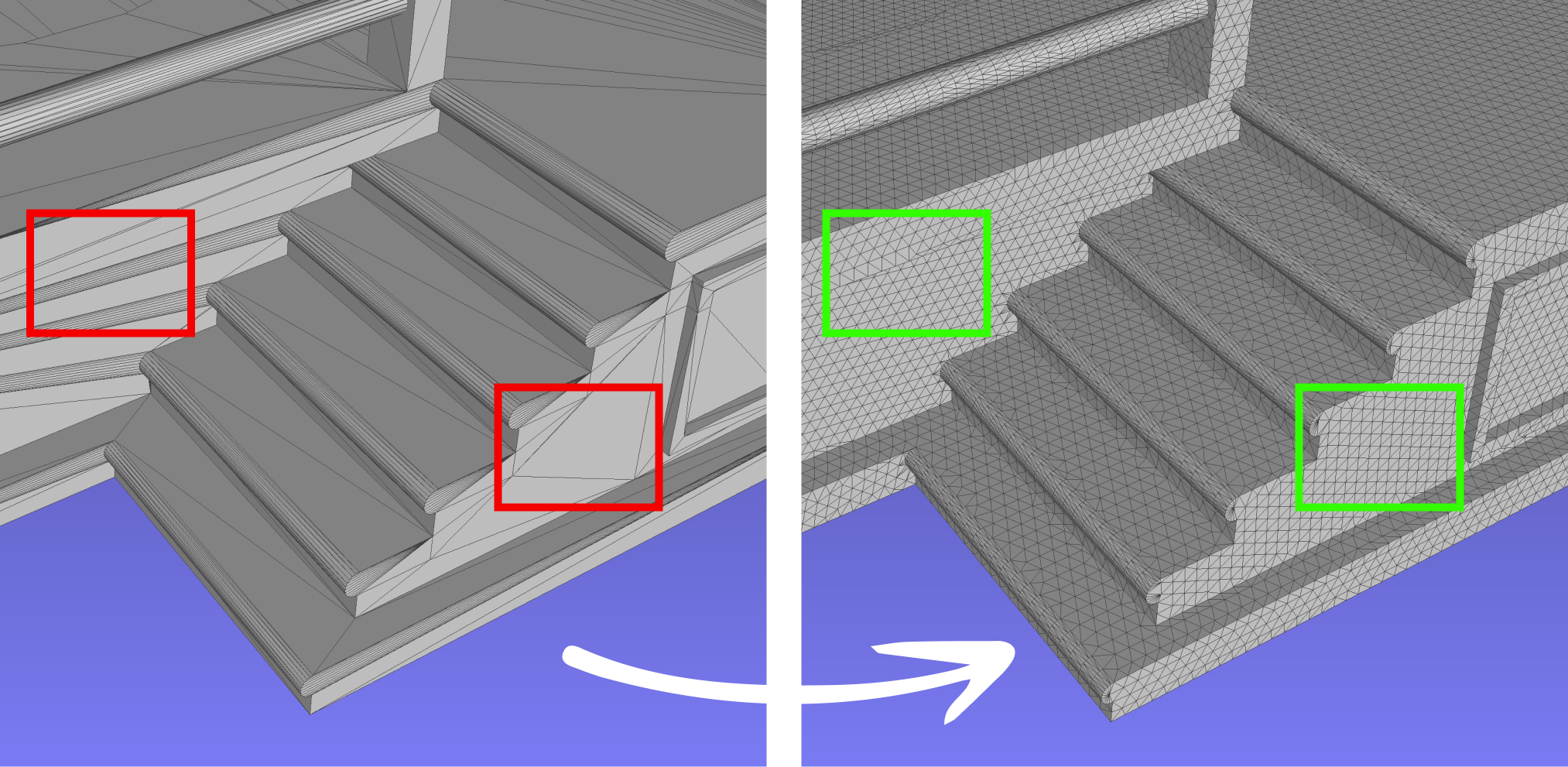}
  \caption{
  \textbf{Exporting NeRF as Mesh.}
  Given the input mesh and the painted NeRF, we remesh the input almost isotropically with planarity constraints and sample vertex colors from the NeRF.
  This technique does not require a UV texture map for the input geometry.
  }
  \label{fig:remesher}
\end{figure}

Each model has a list of associated categories attached to it. 
We compute frequencies of all categories in the entire dataset and assign each model a primary category.
These primary categories are used for both qualitative and quantitative studies.
We demonstrate high-quality painting results on a select set of categories, including electronics, animals, and game characters, in Fig.~\ref{fig:centerpiece}.
See Figs.~\ref{fig:comparison:big},~\ref{fig:shapenetsem:youtube} for more results. 

\paragraph{Comparison with Other Methods}
We compare our method quantitatively with two recent mesh texturing methods, Latent-Paint~\cite{metzer2022latent} and TEXTure~\cite{richardson2023texture} (Fig.~\ref{fig:comparison:big}).
We ran both pipelines on the ShapeNetSem~\cite{chang2015shapenet} dataset using the same 360-degree camera views and text prompts. 
While the TEXTure method handles well-defined camera trajectories, Latent-Paint requires way more views to perform decently; otherwise, we kept their default settings and ensured alignment of the cameras.
We rendered interpolated views of the output models and compared the results of the two pipelines.

To facilitate the quantitative study, we additionally generated painting results for the evaluation views using only the Stable Diffusion \cite{Rombach_2022_CVPR} depth-to-image model.
Although this set of images completely lacks 3D consistency, it provides a useful upper bound on the image fidelity that is attainable with the generative model. 

After processing all models with the selected methods, we render their 360-degree spin views using synchronized camera setups and aggregate them in the video gallery (Fig.~\ref{fig:shapenetsem:youtube}).

A closer look at the output renders in the video (also the car model in Fig.~\ref{fig:comparison:zoomin}) reveals discernible quality differences between different geometry painting methods. 
We can see that the original ShapeNet~\cite{chang2015shapenet} textures are rather primitive. 
Latent-Paint~\cite{metzer2022latent} exhibits blurred and overall coarse texturing. 
TEXTure~\cite{richardson2023texture} produces much more realism and details; however, compared to our method, its output contains spurious artifacts and texture filtering issues. 
This effect is prevalent in complex meshes containing many fine-grain geometry details. 
We observe that both prior methods have distinct artifacts that stem from the effective resolution of the UV texture maps, texture atlas patch discontinuities, and imperfect UV unwrapping.
These issues are further exacerbated when differentiable rendering is employed.
Our method is free of these issues; refer to the Appendix for discussion.

\paragraph{Compute Requirements}
Unlike the other two methods, whose memory footprint fluctuates depending on the 3D model complexity and requires at least 16GB GPU RAM, our method's resources are defined purely by NeRF configuration and are fixed across the whole dataset to 12GB RAM.
Our pipeline configured as stated above takes $\sim$15-20 min to complete, which is on par with the competition.

\paragraph{Quantitative Evaluation}
We execute our pipeline, collect the output NeRF, and sample it at 8 different evaluation views at $45^\circ$ increments.
Using collections of these views obtained for all models in the dataset, 
we compared distribution metrics between each method and the reference (no 3D consistency) for the whole dataset and several primary categories. 
Through this evaluation, we aim to understand how close we can get to the upper bound of lifting the learned generative prior in 3D while maintaining 3D consistency by design.
Frechet Inception Distance (FID)~\cite{heusel2017gans} and Kernel Inception Distance (KID)~\cite{kidmmdgan} are the standard metrics for comparing distributions of images: natural or sampled from generative models.

\begin{table*}[t]
    \centering
    \caption{
        Comparison of geometry painting with various methods on ShapeNetSem~\cite{chang2015shapenet} dataset measured through Frechet Inception Distance (FID~$\downarrow$)~\cite{heusel2017gans} metric with various feature extractors. 
        Lower values are better.
        Results with Kernel Inception Distance~\cite{kidmmdgan} metric are in Tab.~\ref{table:exp:kid}.
    }
    \resizebox{\textwidth}{!}{
      \begin{tabular}{%
@{}%
p{0.1\linewidth}%
p{0.2\linewidth}%
x{0.075\linewidth}%
x{0.075\linewidth}%
x{0.075\linewidth}%
x{0.075\linewidth}%
x{0.075\linewidth}%
x{0.075\linewidth}%
x{0.075\linewidth}%
x{0.075\linewidth}%
x{0.075\linewidth}%
x{0.075\linewidth}%
x{0.075\linewidth}%
x{0.075\linewidth}%
@{}%
}

\toprule

\begin{minipage}{\linewidth}
  FID~$\downarrow$~\cite{heusel2017gans} \\
  Features
\end{minipage}
&
Methods & 
\begin{minipage}{\linewidth}
  \centering
  All (11992)\\
  \fixedvspace{0.2\baselineskip}
  \resizebox{\linewidth}{!}{
    \includegraphics{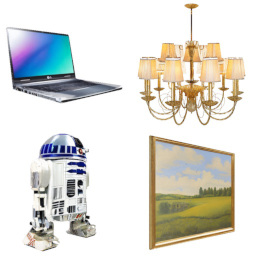}
  }
\end{minipage}
& 
\begin{minipage}{\linewidth}
  \centering
  Misc. (2912)\\
  \fixedvspace{0.2\baselineskip}
  \resizebox{\linewidth}{!}{
    \includegraphics{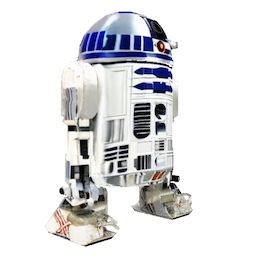}
  }
\end{minipage}
& 
\begin{minipage}{\linewidth}
  \centering
  Chair (682)\\
  \fixedvspace{0.2\baselineskip}
  \resizebox{\linewidth}{!}{
    \includegraphics{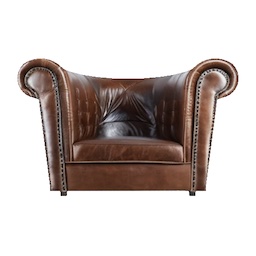}
  }
\end{minipage}
& 
\begin{minipage}{\linewidth}
  \centering
  Lamp (655)\\
  \fixedvspace{0.2\baselineskip}
  \resizebox{\linewidth}{!}{
    \includegraphics{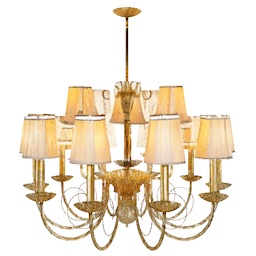}
  }
\end{minipage}
& 
\begin{minipage}{\linewidth}
  \centering
  ChstDrw. (503)\\
  \fixedvspace{0.2\baselineskip}
  \resizebox{\linewidth}{!}{
    \includegraphics{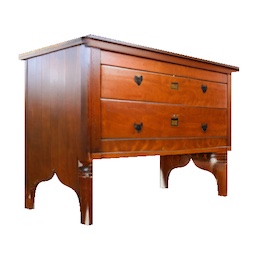}
  }
\end{minipage}
& 
\begin{minipage}{\linewidth}
  \centering
  Table (416)\\
  \fixedvspace{0.2\baselineskip}
  \resizebox{\linewidth}{!}{
    \includegraphics{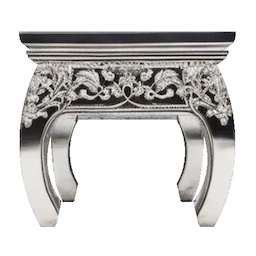}
  }
\end{minipage}
& 
\begin{minipage}{\linewidth}
  \centering
  Couch (405)\\
  \fixedvspace{0.2\baselineskip}
  \resizebox{\linewidth}{!}{
    \includegraphics{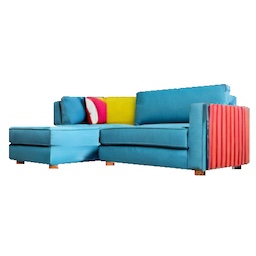}
  }
\end{minipage}
& 
\begin{minipage}{\linewidth}
  \centering
  Computer (241)\\
  \fixedvspace{0.2\baselineskip}
  \resizebox{\linewidth}{!}{
    \includegraphics{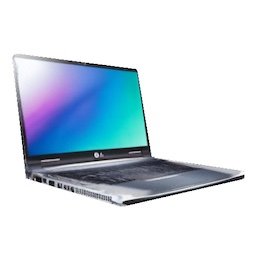}
  }
\end{minipage}
& 
\begin{minipage}{\linewidth}
  \centering
  TV (229)\\
  \fixedvspace{0.2\baselineskip}
  \resizebox{\linewidth}{!}{
    \includegraphics{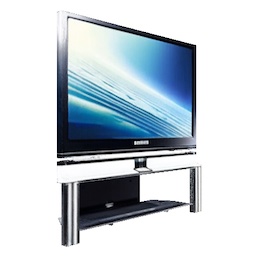}
  }
\end{minipage}
& 
\begin{minipage}{\linewidth}
  \centering
  WallArt (220)\\
  \fixedvspace{0.2\baselineskip}
  \resizebox{\linewidth}{!}{
    \includegraphics{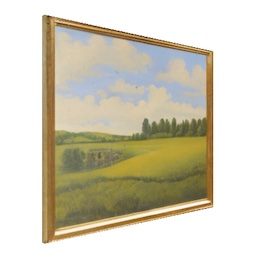}
  }
\end{minipage}
& 
\begin{minipage}{\linewidth}
  \centering
  Bed (218)\\
  \fixedvspace{0.2\baselineskip}
  \resizebox{\linewidth}{!}{
    \includegraphics{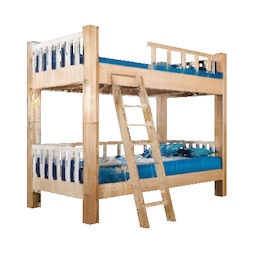}
  }
\end{minipage}
& 
\begin{minipage}{\linewidth}
  \centering
  Cabt. (216)\\
  \fixedvspace{0.2\baselineskip}
  \resizebox{\linewidth}{!}{
    \includegraphics{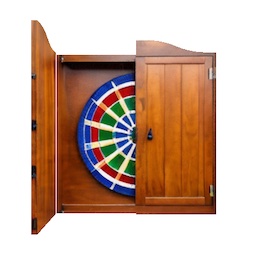}
  }
\end{minipage}
\\

\midrule

\multirow{4}{*}{
\begin{minipage}{\linewidth}
  Inception \\
  \cite{szegedy2016rethinking, heusel2017gans}
\end{minipage}
} &
Orig. texture~\cite{chang2015shapenet}  & 30.10 & 31.82 & 40.79 & 47.61 & 113.6 & 49.18 & 81.39 & 63.02 & 60.89 & 64.92 & 67.37 & 111.9 \\
& Latent-Paint~\cite{metzer2022latent}  & 27.73 & 30.87 & 36.65 & 38.36 & 67.60 & 28.44 & 65.98 & 68.85 & 67.85 & 90.99 & 49.04 & 73.60 \\
& TEXTure~\cite{richardson2023texture}  & 16.10 & 18.34 & 23.44 & 30.75 & 32.65 & 34.98 & 40.40 & 46.48 & 45.85 & 61.23 & 43.04 & 38.88 \\
& \textbf{Ours}  & \textbf{9.60} & \textbf{11.05} & \textbf{16.30} & \textbf{19.54} & \textbf{32.64} & \textbf{22.01} & \textbf{26.23} & \textbf{39.96} & \textbf{29.60} & \textbf{35.77} & \textbf{33.13} & \textbf{36.28} \\

\midrule

\multirow{4}{*}{
\begin{minipage}{\linewidth}
  CLIP \\
  \cite{radford2021learning, clipfid}
\end{minipage}
} &
Orig. texture~\cite{chang2015shapenet}  & 18.86 & 18.71 & 24.89 & 27.66 & 40.15 & 25.72 & 33.57 & 20.60 & 27.29 & 18.86 & 28.79 & 37.07 \\
& Latent-Paint~\cite{metzer2022latent}  & 15.84 & 16.42 & 17.08 & 12.29 & 29.51 & 11.34 & 22.22 & 24.50 & 22.47 & 27.30 & 19.35 & 27.83 \\
& TEXTure~\cite{richardson2023texture}  & 6.85 & 6.85 & 9.62 & 9.37 & 11.29 & 11.00 & 9.48 & 11.28 & 11.38 & 13.17 & 11.09 & 9.79 \\
& \textbf{Ours}  & \textbf{3.24} & \textbf{3.33} & \textbf{3.90} & \textbf{3.47} & \textbf{7.77} & \textbf{4.12} & \textbf{4.69} & \textbf{8.22} & \textbf{6.16} & \textbf{6.18} & \textbf{5.54} & \textbf{7.30} \\

\midrule

\multirow{4}{*}{
\begin{minipage}{\linewidth}
  DINOv2 \\
  \cite{oquab2023dinov2}
\end{minipage}
} &
Orig. texture~\cite{chang2015shapenet}  & 588.1 & 585.9 & 620.6 & 787.3 & 1640. & 883.9 & 1265.6 & 767.1 & 999.4 & 857.9 & 946.3 & 1517. \\
& Latent-Paint~\cite{metzer2022latent}  & 332.9 & 366.1 & 285.8 & 329.0 & 696.6 & 280.6 & 556.3 & 673.4 & 773.9 & 866.5 & 533.4 & 765.1 \\
& TEXTure~\cite{richardson2023texture}  & 175.0 & 194.6 & 181.1 & 278.9 & 321.6 & 248.0 & 282.1 & 404.4 & 501.5 & 580.3 & 276.7 & 366.2 \\
& \textbf{Ours}  & \textbf{125.1} & \textbf{136.7} & \textbf{130.8} & \textbf{181.6} & \textbf{299.4} & \textbf{173.1} & \textbf{239.4} & \textbf{383.0} & \textbf{333.5} & \textbf{312.2} & \textbf{226.3} & \textbf{320.2} \\

\bottomrule
\end{tabular}
    }
    \label{table:exp:fid}
\end{table*}

Following the footsteps of~\cite{clipfid}, we report FID$_\mathrm{CLIP}$ with the CLIP feature extractor.
We additionally propose two new metrics: FID$_\mathrm{DINOv2}$ and KID$_\mathrm{DINOv2}$, which utilize the novel self-supervised feature extraction techniques~\cite{oquab2023dinov2}.
Unlike the decade-old Inception backbone and CLIP, which focuses on named entities, DINOv2 is a powerful self-supervised feature extractor trained on natural images.
All metrics are computed through a verified evaluation protocol of \mbox{\texttt{torch-fidelity}}~\cite{obukhov2020torchfidelity}.
The results of this quantitative study are presented in Tab.~\ref{table:exp:fid}.
Evidently, our method achieves state-of-the-art fidelity to the generative prior while maintaining 3D consistency.

\paragraph{Geometry Export}
The output of our pipeline is contained in the final NeRF reconstruction.
While NeRF as a 3D asset format gains popularity as hardware acceleration catches up, we take an extra step to transfer the generated painting back into a standard editable format.
Since we do not require UV texture maps on the input and want to support use cases such as \mbox{Point-E} discussed below, we opt for transferring colors to the input mesh vertices.
However, to ensure sufficient spatial resolution for such a scheme, vertices should be uniformly distributed on the surface of the input, which is usually not the case.
To overcome this issue, we designed an algorithm for approximately-isotropic remeshing~\cite{obukhov2023remesher} that preserves the input geometry and only focuses on planar regions (Fig.~\ref{fig:remesher}).
Using our remeshing technique helps obtain an identical mesh but with sufficient resolution for color transfer.
Thanks to unambiguous color querying from our view-invariant NeRF flavor, we directly transfer color onto the remeshed input by sampling NeRF at all vertices locations.
We further note, that the output asset files with per-vertex colors occupy significant space, which can be reclaimed by compression techniques such as DRACO~\cite{draco2017compression}.

\begin{figure}
\centering
  \includegraphics[width=\linewidth]{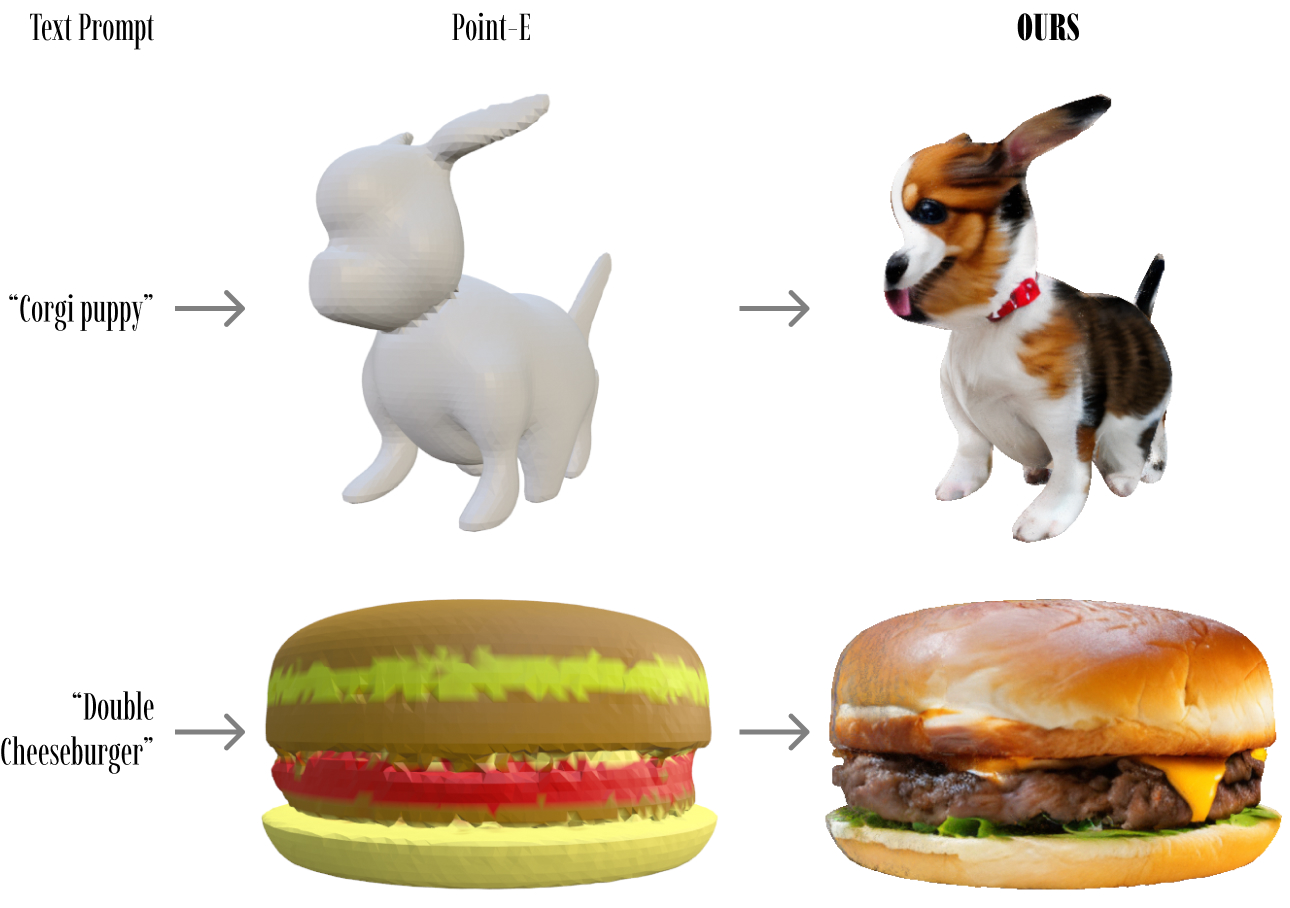}
  \captionof{figure}{
      \textbf{Painting \mbox{Point-E}~\cite{nichol2022point}}. 
      We extend our pipeline to pure text-to-3D by chaining it after \mbox{Point-E}.
      The same text prompt is used to generate the geometry and then repaint it with our method.
  }
  \label{fig:pointe}
\end{figure}

\paragraph{Pure Text-to-3D via Point-E}
We extend our pipeline with \mbox{Point-E}~\cite{nichol2022point}, a diffusion-based generative model that produces 3D point clouds from text prompts. 
Following~\cite{nichol2022point}, we convert the point cloud generated by \mbox{Point-E} to a signed distance field and use marching cubes with grid size $64$ to obtain the mesh serving as an input to our method. 
Since the resulting geometry has surface normals of limited quality, we skip inpainting zoning in our method.
Fig.~\ref{fig:pointe} demonstrates an overall pipeline that takes only a text prompt as the input and outputs a mesh with improved painting. 
From the opposite point of view, since \mbox{Point-E} cannot generate detailed textures, our method can be seen as a downstream modular extension of \mbox{Point-E} to boost the texture quality of the produced 3D models. 

\section{Discussions and Conclusion}
\renewcommand{\thefootnote}{}
\footnotetext{
We thank Shengyu Huang for proofreading this manuscript.
}
\renewcommand{\thefootnote}{\arabic{footnote}}
In this work, we presented a novel pipeline combining a generative 2D diffusion prior and 3D neural radiance fields as \textit{standalone} modules and demonstrated their ability to paint the input geometry using a text prompt in a 3D-consistent manner. 
We conducted a large-scale study on the ShapeNetSem~\cite{chang2015shapenet} dataset and demonstrated the advantages of our approach against several prior art methods on a wide range of object categories. 
We believe that our pipeline will reach the community of artists, content creators, and game developers and enable quick prototyping of 3D assets, particularly from existing ones, thus giving them a new life. 


\clearpage
\renewcommand{\thesection}{\Alph{section}}
\setcounter{section}{0}

\twocolumn[
  \centering
  \Large
  \textbf{Breathing New Life into 3D Assets with Generative Repainting} \\
  \vspace{0.5em}
  Supplementary Material \\
  \vspace{1.5em}
] 

\begin{table*}[ht!]
    \centering
    \caption{
        Comparison of geometry painting with various methods on ShapeNetSem~\cite{chang2015shapenet} dataset measured through Kernel Inception Distance (KID~$\downarrow$)~\cite{kidmmdgan} metric with various feature extractors.  
        Standard deviations are given in small font for all values.
        Lower values are better.
    }
    \resizebox{\textwidth}{!}{
        \begin{tabular}{%
@{}%
p{0.1\linewidth}%
p{0.2\linewidth}%
x{0.075\linewidth}%
x{0.075\linewidth}%
x{0.075\linewidth}%
x{0.075\linewidth}%
x{0.075\linewidth}%
x{0.075\linewidth}%
x{0.075\linewidth}%
x{0.075\linewidth}%
x{0.075\linewidth}%
x{0.075\linewidth}%
x{0.075\linewidth}%
x{0.075\linewidth}%
@{}%
}

\toprule

\begin{minipage}{\linewidth}
  KID~$\downarrow$~\cite{kidmmdgan} \\
  Features \\
  Multiplier
\end{minipage}
&
Methods & 
\begin{minipage}{\linewidth}
  \centering
  All (11992)\\
  \fixedvspace{0.2\baselineskip}
  \resizebox{\linewidth}{!}{
    \includegraphics{imgs/thumbnails/All}
  }
\end{minipage}
& 
\begin{minipage}{\linewidth}
  \centering
  Misc. (2912)\\
  \fixedvspace{0.2\baselineskip}
  \resizebox{\linewidth}{!}{
    \includegraphics{imgs/thumbnails/NoCategory}
  }
\end{minipage}
& 
\begin{minipage}{\linewidth}
  \centering
  Chair (682)\\
  \fixedvspace{0.2\baselineskip}
  \resizebox{\linewidth}{!}{
    \includegraphics{imgs/thumbnails/Chair}
  }
\end{minipage}
& 
\begin{minipage}{\linewidth}
  \centering
  Lamp (655)\\
  \fixedvspace{0.2\baselineskip}
  \resizebox{\linewidth}{!}{
    \includegraphics{imgs/thumbnails/Lamp}
  }
\end{minipage}
& 
\begin{minipage}{\linewidth}
  \centering
  ChstDrw. (503)\\
  \fixedvspace{0.2\baselineskip}
  \resizebox{\linewidth}{!}{
    \includegraphics{imgs/thumbnails/ChestOfDrawers}
  }
\end{minipage}
& 
\begin{minipage}{\linewidth}
  \centering
  Table (416)\\
  \fixedvspace{0.2\baselineskip}
  \resizebox{\linewidth}{!}{
    \includegraphics{imgs/thumbnails/Table}
  }
\end{minipage}
& 
\begin{minipage}{\linewidth}
  \centering
  Couch (405)\\
  \fixedvspace{0.2\baselineskip}
  \resizebox{\linewidth}{!}{
    \includegraphics{imgs/thumbnails/Couch}
  }
\end{minipage}
& 
\begin{minipage}{\linewidth}
  \centering
  Computer (241)\\
  \fixedvspace{0.2\baselineskip}
  \resizebox{\linewidth}{!}{
    \includegraphics{imgs/thumbnails/Computer}
  }
\end{minipage}
& 
\begin{minipage}{\linewidth}
  \centering
  TV (229)\\
  \fixedvspace{0.2\baselineskip}
  \resizebox{\linewidth}{!}{
    \includegraphics{imgs/thumbnails/TV}
  }
\end{minipage}
& 
\begin{minipage}{\linewidth}
  \centering
  WallArt (220)\\
  \fixedvspace{0.2\baselineskip}
  \resizebox{\linewidth}{!}{
    \includegraphics{imgs/thumbnails/WallArt}
  }
\end{minipage}
& 
\begin{minipage}{\linewidth}
  \centering
  Bed (218)\\
  \fixedvspace{0.2\baselineskip}
  \resizebox{\linewidth}{!}{
    \includegraphics{imgs/thumbnails/Bed}
  }
\end{minipage}
& 
\begin{minipage}{\linewidth}
  \centering
  Cabt. (216)\\
  \fixedvspace{0.2\baselineskip}
  \resizebox{\linewidth}{!}{
    \includegraphics{imgs/thumbnails/Cabinet}
  }
\end{minipage}
\\

\midrule

\multirow{4}{*}{
\begin{minipage}{\linewidth}
  Inception \\
  \cite{szegedy2016rethinking, heusel2017gans, kidmmdgan} \\
  $\times 0.01$
\end{minipage}
} 
&
Orig. texture~\cite{chang2015shapenet}  & 1.19\spm 0.04 & 1.18\spm 0.09 & 1.40\spm 0.20 & 2.03\spm 0.38 & 7.89\spm 0.79 & 1.61\spm 0.26 & 4.47\spm 0.56 & 2.76\spm 0.47 & 3.04\spm 0.40 & 1.84\spm 0.52 & 2.72\spm 0.36 & 5.98\spm 0.69 \\
& Latent-Paint~\cite{metzer2022latent}  & 1.31\spm 0.05 & 1.37\spm 0.11 & 2.02\spm 0.25 & 1.95\spm 0.36 & 4.52\spm 0.54 & 1.05\spm 0.21 & 4.26\spm 0.39 & 3.84\spm 0.35 & 4.17\spm 0.34 & 3.81\spm 0.58 & 2.14\spm 0.32 & 4.19\spm 0.47 \\
& TEXTure~\cite{richardson2023texture}  & 0.75\spm 0.04 & 0.71\spm 0.08 & 1.19\spm 0.20 & 1.61\spm 0.35 & \textbf{1.79}\spm 0.38 & 1.97\spm 0.32 & 2.37\spm 0.48 & 2.14\spm 0.33 & 2.36\spm 0.31 & 2.10\spm 0.41 & 1.90\spm 0.32 & 1.54\spm 0.29 \\
& \textbf{Ours}  & \textbf{0.44}\spm 0.03 & \textbf{0.38}\spm 0.06 & \textbf{0.65}\spm 0.18 & \textbf{0.80}\spm 0.24 & 1.88\spm 0.48 & \textbf{0.94}\spm 0.22 & \textbf{1.14}\spm 0.30 & \textbf{1.74}\spm 0.36 & \textbf{1.06}\spm 0.24 & \textbf{0.53}\spm 0.16 & \textbf{1.09}\spm 0.22 & \textbf{1.32}\spm 0.41 \\

\midrule

\multirow{4}{*}{
\begin{minipage}{\linewidth}
  CLIP \\
  \cite{radford2021learning, clipfid} \\
  $\times 0.01$
\end{minipage}
} 
&
Orig. texture~\cite{chang2015shapenet}  & 9.33\spm 0.26 & 8.92\spm 0.50 & 13.1\spm 1.38 & 14.6\spm 1.38 & 21.1\spm 1.71 & 13.0\spm 1.19 & 18.7\spm 1.73 & 8.36\spm 1.12 & 14.3\spm 1.41 & 5.89\spm 1.22 & 14.2\spm 1.32 & 17.7\spm 1.76 \\
& Latent-Paint~\cite{metzer2022latent}  & 7.87\spm 0.18 & 7.87\spm 0.37 & 9.36\spm 0.57 & 6.44\spm 0.67 & 17.1\spm 1.23 & 5.47\spm 0.48 & 13.1\spm 0.85 & 12.1\spm 0.81 & 11.2\spm 0.67 & 10.7\spm 0.99 & 10.5\spm 0.91 & 15.7\spm 1.20 \\
& TEXTure~\cite{richardson2023texture}  & 3.18\spm 0.09 & 3.04\spm 0.21 & 5.12\spm 0.46 & 4.84\spm 0.62 & 5.81\spm 0.67 & 5.67\spm 0.52 & 4.82\spm 0.60 & 4.68\spm 0.43 & 4.63\spm 0.47 & 3.99\spm 0.65 & 5.61\spm 0.50 & 4.40\spm 0.48 \\
& \textbf{Ours}  & \textbf{1.36}\spm 0.06 & \textbf{1.30}\spm 0.13 & \textbf{1.69}\spm 0.32 & \textbf{1.51}\spm 0.27 & \textbf{3.73}\spm 0.70 & \textbf{1.90}\spm 0.34 & \textbf{2.07}\spm 0.39 & \textbf{2.98}\spm 0.61 & \textbf{2.14}\spm 0.48 & \textbf{0.96}\spm 0.27 & \textbf{2.09}\spm 0.36 & \textbf{3.12}\spm 0.62 \\

\midrule

\multirow{4}{*}{
\begin{minipage}{\linewidth}
  DINOv2 \\
  \cite{oquab2023dinov2} \\
  $\times 1.0$
\end{minipage}
} 
&
Orig. texture~\cite{chang2015shapenet}  & 2.40\spm 0.06 & 2.36\spm 0.15 & 3.65\spm 0.37 & 4.95\spm 0.47 & 11.9\spm 0.98 & 5.94\spm 0.61 & 14.1\spm 1.46 & 5.42\spm 0.85 & 9.80\spm 0.96 & 3.75\spm 0.75 & 7.56\spm 0.69 & 9.74\spm 0.79 \\
& Latent-Paint~\cite{metzer2022latent}  & 1.01\spm 0.02 & 1.04\spm 0.05 & 1.85\spm 0.25 & 2.09\spm 0.25 & 5.51\spm 0.51 & 1.62\spm 0.25 & 6.61\spm 0.83 & 5.49\spm 0.54 & 8.87\spm 0.71 & 3.37\spm 0.49 & 4.26\spm 0.50 & 5.21\spm 0.52 \\
& TEXTure~\cite{richardson2023texture}  & 0.53\spm 0.02 & 0.50\spm 0.03 & 1.18\spm 0.24 & 1.76\spm 0.27 & 2.56\spm 0.43 & 1.67\spm 0.27 & 3.32\spm 0.62 & 3.72\spm 0.50 & 3.97\spm 0.61 & 2.17\spm 0.44 & 2.21\spm 0.31 & 2.37\spm 0.39 \\
& \textbf{Ours}  & \textbf{0.38}\spm 0.01 & \textbf{0.35}\spm 0.03 & \textbf{0.63}\spm 0.21 & \textbf{1.07}\spm 0.21 & \textbf{2.01}\spm 0.48 & \textbf{1.03}\spm 0.21 & \textbf{1.90}\spm 0.64 & \textbf{3.14}\spm 0.66 & \textbf{2.24}\spm 0.43 & \textbf{0.86}\spm 0.19 & \textbf{1.08}\spm 0.23 & \textbf{1.58}\spm 0.37 \\

\bottomrule
\end{tabular}
    }
    \label{table:exp:kid}
\end{table*}

\section{Large-Scale Study of ShapeNetSem}
\label{sec:largescale}

Out of 12,288 models in the dataset, we processed 11,992 with all methods. 
The remaining 296 models either had flat geometry or could not be processed by the Latent-Paint~\cite{metzer2022latent} pipeline, TEXTure~\cite{richardson2023texture}, or both.
The failure cases happened most commonly due to the complex geometry not fitting 16GB GPU RAM within the respective method pipeline or failures in \texttt{xatlas} texture UV unwrapping module~\cite{levy2002least}.
Our method produced results consistently even on these models, but for a fair comparison, we excluded these models completely.

In addition to the FID~\cite{heusel2017gans} evaluation from Tab.~\ref{table:exp:fid}, we provide a quantitative evaluation of all pipelines on ShapeNetSem with the KID metric~\cite{kidmmdgan} in Tab.~\ref{table:exp:kid}.

The ability of our method to handle complex geometry, low memory footprint, weak dependence on the geometry format or the rendering pipeline, and potentially unknown texture coordinates -- all these properties make our method a reliable go-to solution for 3D assets revamping.

\section{Subjective User Study}

\begin{figure}[t]
  \centering
  \includegraphics[width=\linewidth, trim={0 8.5em 0 0}, clip]{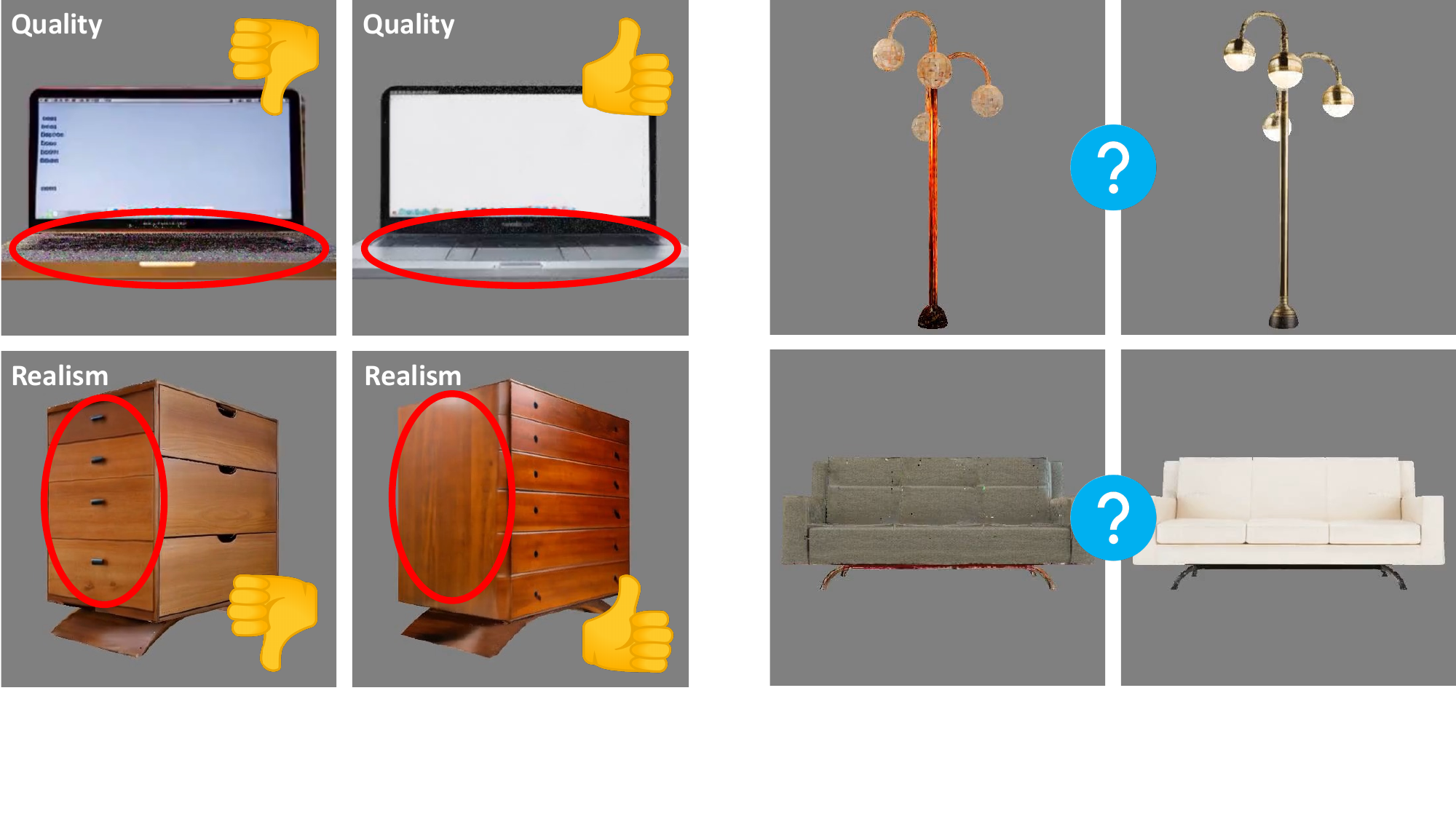}
  \caption{
  \textbf{Subjective Study.}
  Left: User instruction with quality and realism judgment examples;
  \textbf{Right}: Two validation questions ``Which one is better?'' shared among all subjects to ensure engagement (the right column answers were expected for a pass).
  }
  \label{fig:userstudy}
\end{figure}

We conducted a limited crowd-sourced perceptual comparison between Latent-Paint~\cite{metzer2022latent}, TEXTure~\cite{richardson2023texture}, and our method.
The study was based on 50 randomly sampled models from 10 categories, c.f.~Tab.~\ref{table:exp:fid}.
Subjects were instructed (Fig.~\ref{fig:userstudy}, left) to analyze and vote for higher quality and realism after observing a full 360$^\circ$ spin of models painted with a pair of methods, side by side. 
Each subject submitted 20 votes, plus 2 validation questions with predefined correct answers (Fig.~\ref{fig:userstudy}, right).
35 subjects participated in our study, of which 29 (83\%) passed the validation.
638 votes were collected, ensuring at least 3 votes for every pair, and aggregated into preference scores with the Crowd Bradley-Terry~\cite{bradley1952rank} model. 
The resulting scores were (log-scale, up to additive constant, 95\% confidence intervals, higher is better):
$
S_\mathrm{Latent-Paint}=0.15_{\pm 0.15},
S_\mathrm{TEXTure}=0.30_{\pm 0.11},
S_\mathrm{ours}=\mathbf{1.86}_{\pm 0.13}
$.
The scores agree with the quantitative results.

\section{ShapeNet Rendering Settings}

\begin{figure*}[t]
  \centering
  \includegraphics[width=\linewidth]{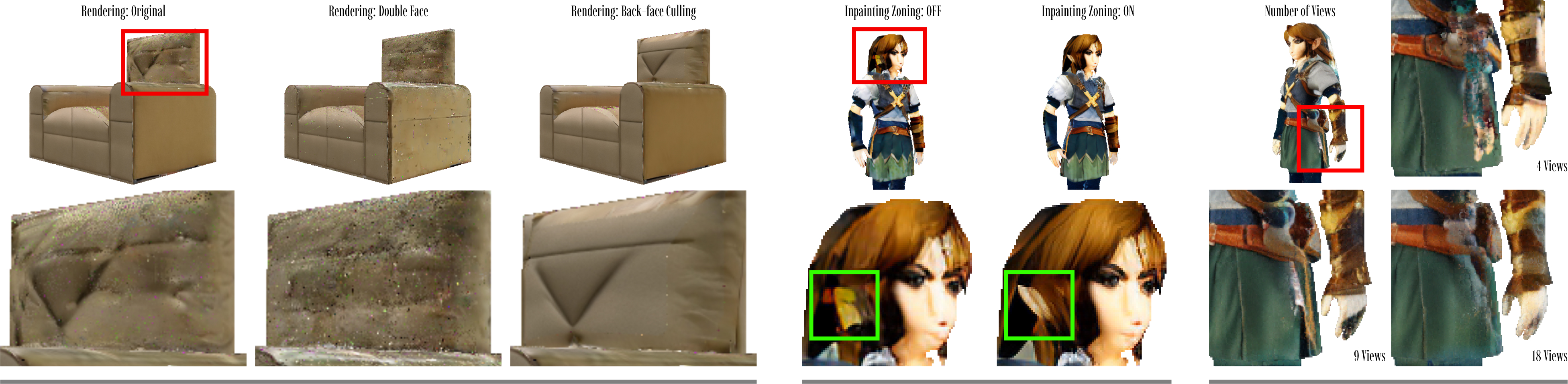}
  \caption{
    \textbf{Ablations.}
    \textbf{Left:} Rendering Settings Comparison with TEXTure~\cite{richardson2023texture} method.
      Employing back-face culling achieves the best result compared with the original%
      \href{https://github.com/TEXTurePaper/TEXTurePaper/issues/8}{$^1$}
      and double-face rendering settings. 
    \textbf{Middle:} Visibility Score Refinement produces more realistic details on surfaces seen at sharp angles, such as the ear.
    \textbf{Right:} Low number of generated views (4) leads to poor coverage of the input geometry, 18 results in over-smoothing, and 9 is a trade-off.
  }
  \label{fig:ablations}
\end{figure*}

To facilitate a fair comparison of different methods on the ShapeNetSem dataset~\cite{chang2015shapenet}, we choose the mesh rendering settings in all pipelines such that the output result is adequate for all methods.
Notably, TEXTure~\cite{richardson2023texture} relies on mesh normals to determine inpainting regions.
However, a subset of ShapeNetSem~\cite{chang2015shapenet} meshes have faces with inappropriately oriented surface normals. 
For these meshes, directly passing them as input to TEXTure~\cite{richardson2023texture} produces corrupt texturing. 

To address this issue, we utilize back-face culling of mesh to disable the rendering of mesh faces that are oriented away from the camera. 
We build our method on top of PyTorch3D~\cite{ravi2020pytorch3d}, which provides a built-in implementation of back-face culling. 
However, since both Latent-Paint~\cite{metzer2022latent} and TEXTure~\cite{richardson2023texture} pipelines rely on the Kaolin renderer~\cite{KaolinLibrary}, which did not implement back-face culling as of the time of writing, we implemented back-face culling in software.
This allowed us to address the rendering discrepancy and level the settings for all pipelines.

We experimented with double-face rendering as an alternative approach to resolving face orientation issues. 
However, the result of using double-face rendering is worse than that of using back-face culling, as seen in Fig.~\ref{fig:ablations} (left). 
We suspect this is due to areas of the mesh having overlapping front-facing faces in the double-face rendering setting, thereby negatively affecting texture back-projection in the TEXTure~\cite{richardson2023texture} method. 
Overall, our rendering protocol is chosen to maximize the output quality of the pipelines relying on differential rendering under complex geometry. 

\section{Ablation: Inpainting Zoning} 
Inpainting zoning works in areas of the mesh that face away from the camera in one generated view so that they can be further refined in the subsequent views. 
Fig.~\ref{fig:ablations} (middle) shows that our refinement scheme brings more details to the areas of the model with challenging visibility constraints. 

\section{Ablation: Number of Input Views}
We show a qualitative comparison between models painted using various numbers of input views in Fig.~\ref{fig:ablations} (right). 
With just 4 input views, we find holes and artifacts on the object's surface. 
With 18 views, the shape is smooth, but the generated color lacks detail.
The choice of 9 views achieves the best quality.   

\section{Prompt Augmentation}
\label{sec:promptaug}

\begin{figure}
\centering
  \includegraphics[width=\linewidth]{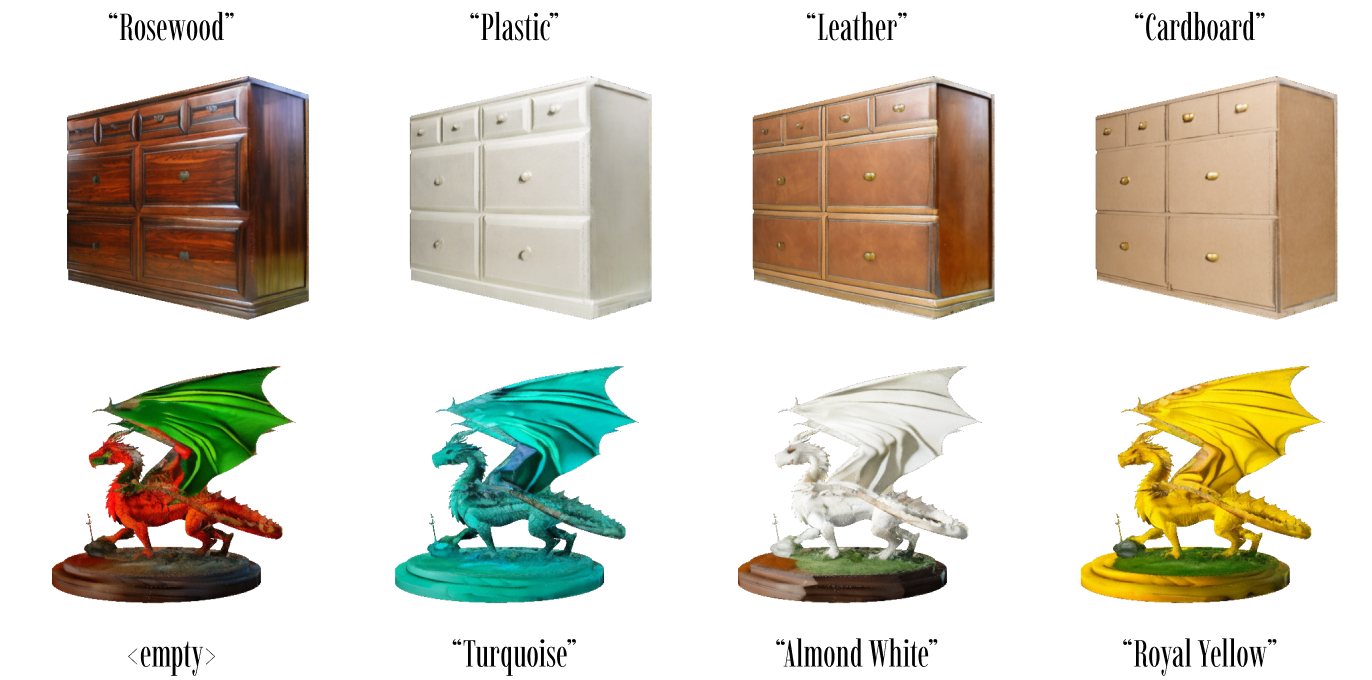}
  \captionof{figure}{
    \textbf{Style Specifier Guidance}. 
      In the given examples, prompts take the following form: 
      ``\textit{A photo of a $\{\!\{$material$\}\!\}$ dresser}'' (top) and ``\textit{A photo of a $\{\!\{$color$\}\!\}$ dragon}'' (bottom).
  }
  \label{fig:custom}
\end{figure}

Our method transparently exposes the style guidance functionality of the underlying generative models.
It permits prompt augmentation, enabling greater variety in the generated painting while preserving 3D consistency. 
Specifically, our pipeline extends the input object description prompt as follows: ``\textit{A photo of a \{\!\{modifier\}\!\} \{\!\{object\}\!\}, \{\!\{\textit{dir}\}\!\} view}''. 
The ``\textit{\{\!\{modifier\}\!\}}'' style specifier term could be the color or the material of the object. 
In the same vein as text guides image generation models, the texture of our 3D models changes according to the modifier, as shown in Fig.~\ref{fig:custom}.

{%
\small
\clearpage
\balance
\bibliographystyle{ieee_fullname}
\bibliography{egbib}
}

\end{document}